**Title:**

Leveraging Support Vector Regression, Radiomics and Dosiomics for Outcome Prediction in Personalized Ultra-fractionated Stereotactic Adaptive Radiotherapy (PULSAR)

**Authors:**

Yajun Yu[1,2], Steve Jiang[1,2], Robert Timmerman[1], and Hao Peng[1,2*]

**Author affiliations:**

[1]Department of Radiation Oncology, University of Texas Southwestern Medical Center, Dallas, TX 75390, USA.

[2]Medical Artificial Intelligence and Automation Laboratory, University of Texas Southwestern Medical Center, Dallas, TX 75390, USA.

*Corresponding author

Email: hao.peng@utsouthwestern.edu

Postal address: Department of Radiation Oncology, University of Texas Southwestern Medical Center, 2280 Inwood Rd., Dallas, TX 75390, USA.

**Abstract**

Personalized ultra-fractionated stereotactic adaptive radiotherapy (PULSAR) is a novel treatment that delivers radiation in pulses of protracted intervals. Accurate prediction of gross tumor volume (GTV) changes through regression models has substantial prognostic value. This study aims to develop a multi-omics based support vector regression (SVR) model for predicting GTV change. A retrospective cohort of 39 patients with 69 brain metastases was analyzed, based on radiomics (MRI images) and dosiomics (dose maps) features. Delta features were computed to capture relative changes between two time points. A feature selection pipeline using least absolute shrinkage and selection operator (Lasso) algorithm with weight- or frequency-based ranking criterion was implemented. SVR models with various kernels were evaluated using the coefficient of determination ($R^2$) and relative root mean square error (RRMSE). Five-fold cross-validation with 10 repeats was employed to mitigate the limitation of small data size. Multi-omics models that integrate radiomics, dosiomics, and their delta counterparts outperform individual-omics models. Delta-radiomic features play a critical role in enhancing prediction accuracy relative to features at single time points. The top-performing model achieves an $R^2$ of 0.743 and an RRMSE of 0.022. The proposed multi-omics SVR model shows promising performance in predicting continuous change of GTV. It provides a more quantitative and personalized approach to assist patient selection and treatment adjustment in PULSAR.

## 1. Introduction

Recently, our institution advocated a new radiation treatment paradigm, i.e. personalized ultra-fractionated stereotactic adaptive radiotherapy (PULSAR)(Moore *et al.*, 2021), which attempts to administer tumoricidal doses in a pulse mode at intervals of two to four weeks, rather than daily or every other day delivery in conventionally fractionated radiation therapy (CFRT)(Slawson *et al.*, 1988; Demaria *et al.*, 2021) or stereotactic body radiation therapy (SBRT)(Fakiris *et al.*, 2009; Lo *et al.*, 2010). The extended interval of fractions in PULSAR allows more normal tissue recovery after injury, and meanwhile provides sufficient time for the tumor and its microenvironment to undergo changes (e.g. tumor size, tumor shape, and biomarker expression)(Moore *et al.*, 2021; Miljanic *et al.*, 2022; Peng *et al.*, 2024a; Peng *et al.*, 2024b). Gross tumor volume (GTV) is a critical indicator of treatment response during and after therapy(Burnet *et al.*, 2004). To explore the therapeutic efficacy of PULSAR, it is desirable to predict the dynamics of GTV as early as possible during the treatment. In our previous study, we have studied the GTV change as a binary classification task, leveraging pre-treatment and intra-treatment magnetic resonance images (MRIs) alongside dose distributions(Zhang *et al.*, 2024; Dohopolski *et al.*, 2024; Yu *et al.*, 2025). We developed a multi-omics based outcome prediction model to classify between the response and non-response groups with a threshold of 20% GTV reduction at 3-month follow-up(Zhang *et al.*, 2024).

However, a regression model provides more detailed and quantitative insights compared to a classification model when predicting tumor changes. There are several benefits of using a regression model over a classification model. First, tumor progression is a continuous process, and a regression model can capture subtle differences in growth or shrinkage that a classification model might miss. Instead of a binary output, the regression model is able to estimate *how much* the tumor changes (e.g., percentage increase or decrease in size), which can be more informative for treatment decisions. Second, physicians often need precise measurements rather than categorical labels. Knowing the extent of tumor change can help tailor treatment plans, such as adjusting radiation doses, to avoid either under-treatment or over-treatment. Third, using the regression model, a clinician could define specific thresholds (e.g., >20% increase is "progression," 10-20% is "stable disease"), allowing for more customized classification rather than a fixed categorical output. Fourth, if we are tracking tumor changes over multiple time points, a regression model may allow us to better quantify trends rather than just assigning discrete labels at each step.

Radiomics is a high-through conversion of medical images into quantitative handcrafted features(Kumar *et al.*, 2012; Lambin *et al.*, 2017), while dosiomics further integrates the characterization of the spatial heterogeneity of dose distributions(Liang *et al.*, 2019; Zhang *et al.*, 2023). Their combination have been broadly applied in classification model construction to predict discrete categories, such as tumor prognosis(Elshafeey *et al.*, 2019; Wang *et al.*, 2023; Jiang *et al.*, 2023), treatment selection(Teruel *et al.*, 2014; Mu *et al.*, 2020; Wang *et al.*, 2024), and treatment response assessment(Nasief *et al.*, 2019; Murakami *et al.*, 2022; Zhang *et al.*, 2023; Zhang *et al.*, 2024). However, their application in regression tasks remains limited. One study quite similar to our approach developed a machine learning regression model that integrates radiomic and dosiomic features to predict absorbed doses in organs at risk, investigating how image texture and dose distribution patterns correlate with actual radiation absorption aiming to facilitate planning for neuroendocrine tumors treated with $^{177}$Lu-DOTATATE therapy(Plachouris *et al.*, 2023).

In this work, a cohort of brain metastases (BM) patients treated with PULSAR was chosen as a case study. Radiomics and dosiomics derived from MRI scans and dose maps, respectively, were utilized to develop support vector regression (SVR) models for GTV estimation after PULSAR treatment. A comparative analysis of regression models constructed with individual-omics (radiomics, dosiomics or their delta modes) and multi-omics was conducted, aiming to improve predictive accuracy.

## 2. Methods

### 2.1. Study Design

The workflow of this study involves four parts, as displayed in **Figure 1**. In the first part, radiomic features ($\mathbf{R}_{init}$ and $\mathbf{R}_{intra}$) were extracted from initial and intra-treatment MRI scans within the GTV, and the corresponding dose maps were utilized to derive dosiomic features ($\mathbf{D}_{init}$ and $\mathbf{D}_{intra}$). Meanwhile, delta-radiomic ($\mathbf{R}_\Delta$) or delta-dosiomic ($\mathbf{D}_\Delta$) features were computed to represent the change between the initial and intra-treatment features. In the second part, the feature selection was conducted to remove irrelevant and redundant features, using the variance threshold method and linear correlation analysis. To handle with the limitation of small sample size, an ensemble approach with 10 repeats, 5-fold cross validation was selected. In each iteration, least absolute shrinkage and selection operator (Lasso) algorithm(Tibshirani, 1996) was used to identify top features with large weights. Two criteria were applied (see details in Methods section), *X-cnt* and *X-abs.* In the third part, SVR algorithms with four kernels (i.e. linear, radial basis function (RBF), polynomial and sigmoid) were employed to build regression models. In the last part, two metrics were chosen to assess the regression performance, coefficient of determination ($R^2$) and relative root mean square error (RRMSE).

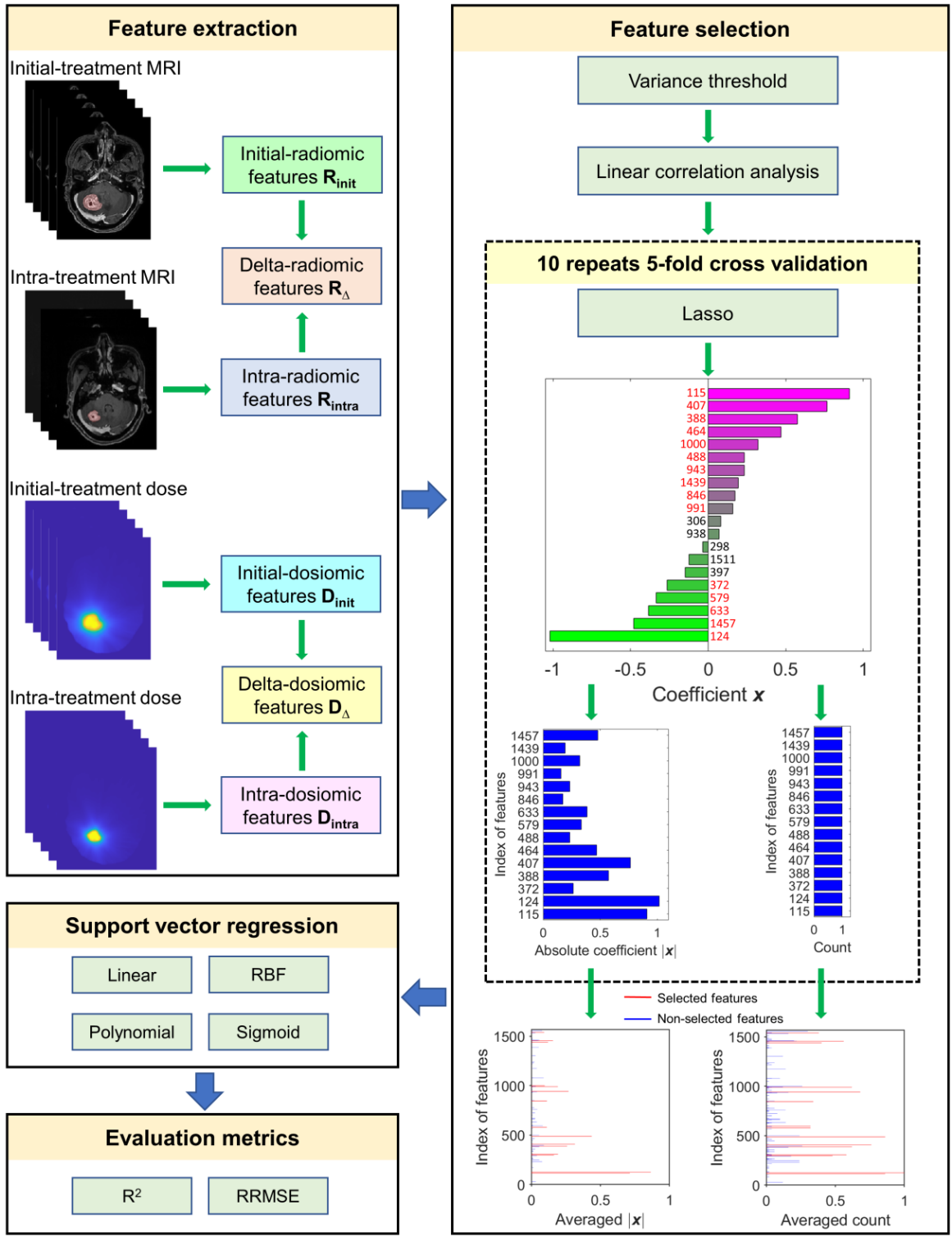

**Figure 1.** Overview of the study design comprising four modules: feature extraction, feature selection, support vector regression, and evaluation metrics. MRI: magnetic resonance image; Lasso: least absolute shrinkage and selection operator; RBF: radial basis function; $R^2$: coefficient of determination; RRMSE: relative root mean square error.

## 2.2. PULSAR treatment and patient data

This was a retrospective study approved by the Institutional Review Board at UT Southwestern Medical Center. Patients with brain metastases were treated with PULSAR using Gamma Knife radiosurgery between November 2021 and May 2023. During PULSAR treatment process, a patient initially underwent a pre-treatment MRI scan before a first treatment course. After 2 to 4 weeks, a second course was delivered according to an intra-treatment MRI scan, with adjustments based on tumor volume changes or presence of vasogenic edema. All MRI images were acquired through axial MRI sequences with T1-weighted enhancement, guaranteeing a consistent and standardized basis for radiomic analysis. Comprehensive initial and intra-treatment data collected from patients undergoing PULSAR encompassed MRI images and dose maps. In addition, follow-up MRI images were acquired to assess tumor volumes 3 months after PULSAR treatment. Lesions were manually identified and contoured to generate GTV masks. Overall, thirty-nine patients diagnosed with single or multiple brain metastases were recruited for PULSAR treatment. This cohort included 69 brain metastasis lesions adopted for analysis.

## 2.3. Feature extraction

To ensure consistency and accuracy in feature extraction, 3D MRI images, 3D dose maps and 3D GTV masks were co-registered and subsequently interpolated to standardize voxel sizes to $1 \times 1 \times 1$ mm$^3$. For each lesion sample at a given treatment stage, 1565 radiomic and 1565 dosiomic features were extracted by PyRadiomics(Van Griethuysen *et al.*, 2017), an open-source Python package compliant with the Image Biomarker Standardization Initiative (IBSI) guidelines(Zwanenburg *et al.*, 2016). The extracted features were categorized into original groups (first-order statistics, shape descriptor, gray level co-occurrence matrix [GLCM], gray level size zone matrix [GLSZM], gray level run length matrix [GLRLM], neighboring gray tone difference matrix [NGTDM], and gray level dependence matrix [GLDM]) and filter-derived groups (filters including Laplacian of Gaussian, wavelet, square, square root, logarithm, exponential, and gradient). To make all feature vectors comparable, each feature vector was standardized and normalized.

Delta-radiomic ($\mathbf{R}_\Delta$) or Delta-dosiomic ($\mathbf{D}_\Delta$) features were calculated based on the relative change between initial features ($\mathbf{R}_{init}$ or $\mathbf{D}_{init}$) and intra-treatment features ($\mathbf{R}_{intra}$ or $\mathbf{D}_{intra}$), formulated as (take $\mathbf{R}_\Delta$ as an example):

$$R_\Delta = \frac{R_{init} - R_{intra}}{R_{init}} \quad (1)$$

### 2.4. Feature selection:

Nine scenarios of feature sets were considered in this study, including individual-omics (i.e. $R_{init}$, $R_{intra}$, $R_\Delta$, $D_{init}$, $D_{intra}$, and $D_\Delta$) and multi-omics (i.e. $R_{init}+R_{intra}+R_\Delta$, $D_{init}+D_{intra}+D_\Delta$, and $R_{init}+R_{intra}+R_\Delta+D_{init}+D_{intra}+D_\Delta$). Owing to the high dimensionality of each feature set (over 1000) with respect to the small data size (69 samples), most features should be eliminated to avoid overfitting. A mixture of feature selection strategies was employed. First, the features with low variance (<0.001) were excluded. Second, Pearson linear correlation analysis was performed to discard any redundant feature with high absolute correlation coefficient (>0.95) to any of its previous features. Third, to mitigate the potential overfitting and bias due to the small data size, a 5-fold cross-validation procedure with 10 repeats was executed for further feature selection. During each repeat of 5-fold cross-validation, the dataset was randomly shuffled and split into five parts. In each iteration, one of the parts was reserved as the test set and the remaining four parts formed the training set. The training set in each iteration was then imported into the Lasso algorithm. Subsequently a fixed number of 20 non-zero coefficients were determined by tuning the regularization parameter, leading to a reservation of 20 corresponding feature candidates. Two criteria for subsequent feature selection are described below:

*X-abs*: The top fifteen maximum absolute values from 20 non-zero coefficients of *x* were reserved in each iteration. The recorded absolute coefficients were averaged across all iterations to determine the feature ranking.

*X-cnt*: Unlike the *X-abs* criterion, the top fifteen features with maximum absolute coefficients from 20 candidates were counted once per iteration, followed by frequency-based ranking across whole iterations.

For each criterion, 15 features were ultimately selected and ranked based on their importance scores averaged through 50 iterations.

### 2.5. Support vector regression:

Support vector regression (SVR) is an adaption of support vector machine (SVM) to transform input features into high-dimensional spaces to find a continuous-valued function accompanied with an $\varepsilon$-insensitive tube region, while training samples that lie outside the tube region(Smola and Schölkopf, 2004; Awad and Khanna, 2015). The transformation of feature spaces can be realized by kernel trick. Four kernel functions were used in our SVR models, including linear, RBF, polynomial, and sigmoid. The hyperparameters in terms of each kernel are optimized through the grid-search and 5-fold cross-validation. More details for the hyperparameters tuning can be found in **Supplementary Material**.

**2.6. Statistical analysis:**

Two evaluation metrics were adopted to assess the regression model performance: the coefficient of determination ($R^2$) and the relative root mean square error (RRMSE). The definitions are as follows:

$$R^2 = 1 - \frac{\sum_{i=1}^{n}(y_i - \hat{y}_i)^2}{\sum_{i=1}^{n}(y_i - \bar{y})^2} \quad (2)$$

$$\text{RRMSE} = \sqrt{\frac{\frac{1}{n}\sum_{i=1}^{n}(y_i - \hat{y}_i)^2}{\sum_{i=1}^{n}(\hat{y}_i)^2}} \quad (3)$$

where $y_i$ is the actual value of the *i*-th sample, $\hat{y}_i$ is the predicted value of the *i*-th sample, $\bar{y}$ is the mean of all actual values, and *n* is the number of samples. $R^2$ measures the proportion of the variable for a dependent variable that is predictable from the independent variable in a regression model, with values closer to 1 indicating better performance. RRMSE is a dimensionless, normalized metric that quantifies the difference between the predicted and actual values, with lower values reflecting better predictive accuracy.

To minimize bias in prediction accuracy as well as identifying an empirical estimate of performance variability, a five-fold cross validation with ten repeats was employed in the regression model construction. This approach resulted in 50 quantitative values for each evaluation metric per model. Model performance evaluation was conducted by averaging these values, and the results were reported alongside the corresponding 95% confidence intervals (CIs). Linear correlation coefficient between two variables was calculated to evaluate the degree of their relationship. Variance inflation factor (VIF) was used to examine the multicollinearity issue in a regression model(Thompson *et al.*, 2017). Cohen's *d* was computed to express the effect size between two groups(Goulet-Pelletier and Cousineau, 2018). All analyses were conducted using MATLAB R2023b and Python v3.9.18.

**3. Results**

### 3.1. Patient Characteristics

A total of thirty-nine patients were included in this study, comprising 14 males and 25 females with a median age of 61 years (range: 28 to 84 years). This cohort included an initial set of 69 BM lesions, of which 26 were single metastases and 43 were multiple metastases. The GTVs varied between 14.5 mm$^3$ and 37607.6 mm$^3$.

### 3.2. Selected Features

For each MRI image or dose map, a total of 1565 features were first extracted by the PyRadiomics package(Van Griethuysen *et al.*, 2017). A summary of the selected features is presented in **Supplementary Dataset**. After the feature selection phase, 15 features were retained and ranked for each of the nine scenarios using either the *X-abs* or *X-cnt* criterion. The final selected feature sets using the *X-cnt* criterion are presented in **Figure 2**, while the results for *X-abs* criterion are exhibited in **Figure S1** (**Supplementary Material**).

**Figure 2.** Selected 15-feature sets of nine scenarios for *X-cnt* criterion. The background colors represent different scenarios. The number in each bracket after a feature name refers to the feature index in the originally extracted 1565 features. New emerging features, not identified in the upstream analysis, are marked with red diamonds.

Among the selected features, most are from wavelet filter-derived groups. Comparing 15-feature sets selected by both criteria in each scenario, a majority of features indeed overlap (**Figure 2** and **Figure S1**), although their rankings slightly differ. Notably, the shared features are more likely to cluster in the top-ranked positions. Furthermore, for the multi-omic scenario, delta features constitute the majority. The feature set of **$R_{init}$+$R_{intra}$+$R_\Delta$+$D_{init}$+$D_{intra}$+$D_\Delta$** is mainly composed of delta-radiomic features (#1-4, 6, 7). It is worth noting that all the multi-omic features selected by *X-abs* criterion are derived from the upstream feature sets (**Figure S1**), while under *X-cnt* criterion, some new features emerge in the multi-omic scenario (#12, 15, marked by red diamonds in **Figure 2**).

### 3.3. Regression Model Performance

The SVR performance for GTV prediction with 10 repeats, 5-fold cross-validation evaluated by $R^2$ and RRMSE, is shown in **Figure 3** and **4** (with RBF kernel), and **Figure S2-S7** (with linear, polynomial and sigmoid kernels, see **Supplementary Material**). For the convenience of comparison, the coordinate ranges for $R^2$ and RRMSE are restricted to [-0.2, 0.8] and [0, 0.2], respectively. The performance metrics from the *X-abs* criterion across all four kernels, as well as the *X-cnt* criterion with the polynomial kernel, show considerable variability—particularly in $R^2$ values—highlighting the inferior performance of the *X-abs* criterion and the polynomial kernel for GTV prediction. In contrast, the regression performance using the *X-cnt* criterion with RBF (**Figure 4**), linear, and sigmoid kernels exhibits consistent trends, with both $R^2$ and RRMSE improving as the number of features increases.

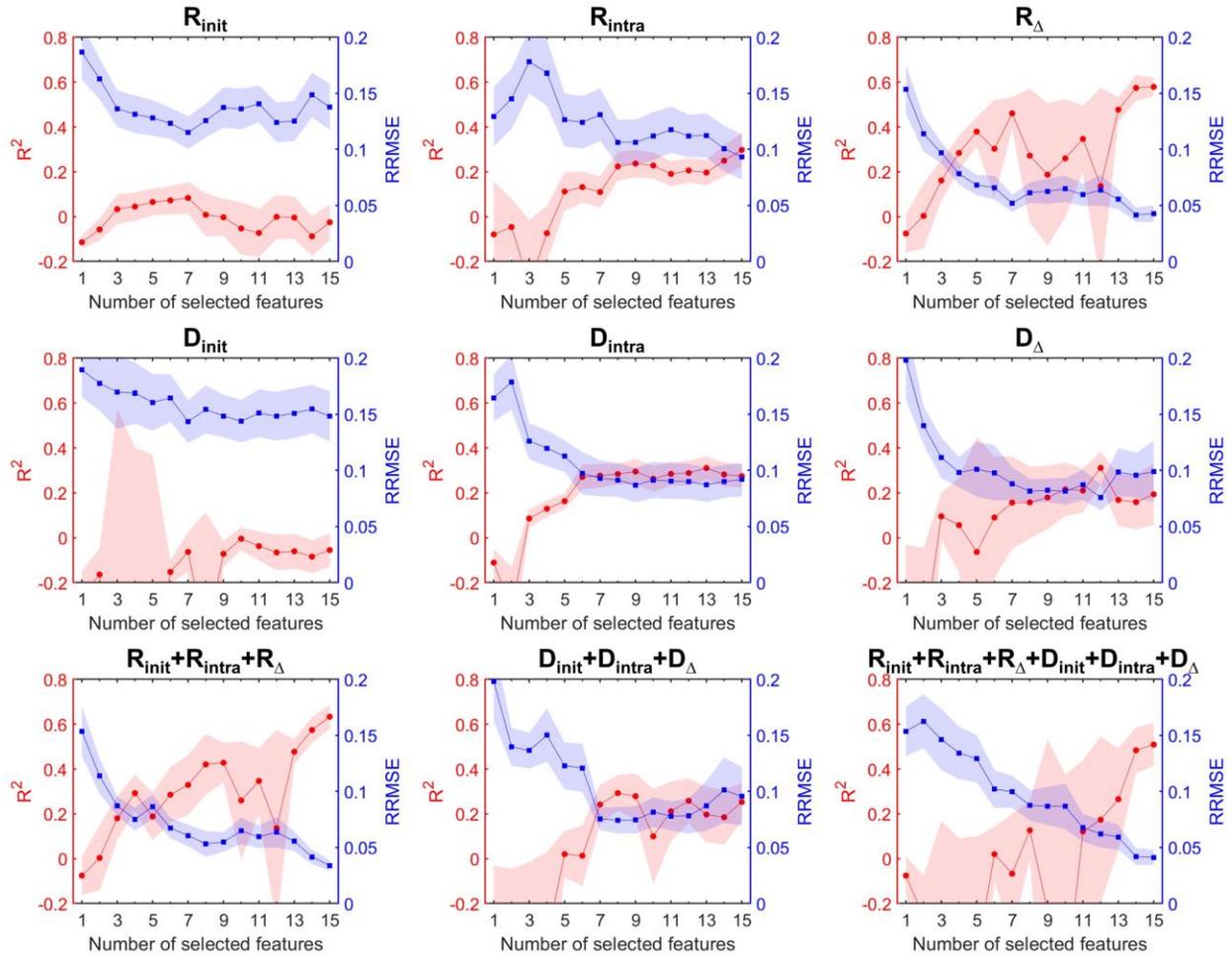

**Figure 3.** Regression evaluation metrics ($R^2$ and RRMSE) for nine scenarios as a function of the number of selected features (kernel: RBF). The *X-abs* criterion is applied in feature selection. The shaded areas represent 95% CIs.

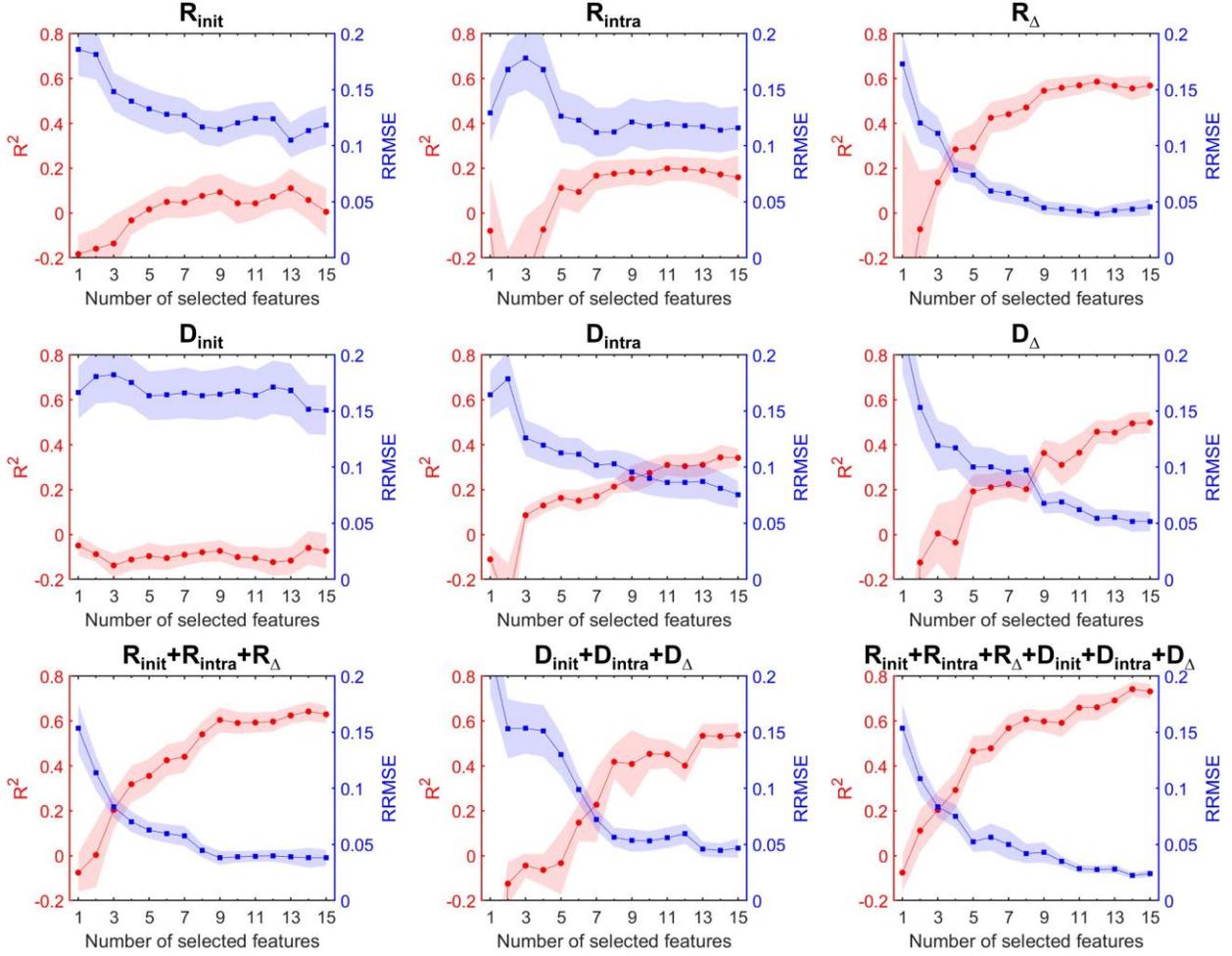

**Figure 4.** Regression evaluation metrics ($R^2$ and RRMSE) for nine scenarios as a function of the number of selected features (kernel: RBF). The *X-cnt* criterion is applied in feature selection. The shaded areas represent 95% CIs.

The best performance across all the numbers of selected features in each scenario (*X-cnt* criterion) is summarized in **Table 1** and **2**, with the highest $R^2$ and lowest RRMSE across four kernel functions highlighted in bold. The superior values are predominantly found in the RBF kernel columns, particular in the multi-omic scenarios (except for $R^2$ at $D_{init}+D_{intra}+D_\Delta$). Overall, the multi-omic prediction models outperform the upstream individual models, in terms of both $R^2$ and RRMSE. The combination of radiomics, dosiomics and the delta features ($R_{init}+R_{intra}+R_\Delta+D_{init}+D_{intra}+D_\Delta$) with RBF kernel achieves the best performance, with $R^2 = 0.743$ and RRMSE = 0.022. **Figure 5** shows scatter plots of actual GTVs versus predicted GTVs for nine scenarios based on the *X-cnt* criterion. The scatter plots further confirm that the multi-omic regression model yields more accurate predictions compared to others.

**Table 1.** Best $R^2$ results for each scenario in the *X-cnt* criterion. The highest $R^2$ across four kernels is highlighted in bold. 95% CIs are included in brackets.

|  | Linear | | RBF | | Polynomial | | Sigmoid | |
|---|---|---|---|---|---|---|---|---|
|  | Best # of selected features | $R^2$ (95% CI) | Best # of selected features | $R^2$ (95% CI) | Best # of selected features | $R^2$ (95% CI) | Best # of selected features | $R^2$ (95% CI) |
| $R_{init}$ | 6 | 0.073 (0.010-0.135) | 13 | **0.111** (0.023-0.199) | 3 | 0.075 (-0.005-0.155) | 6 | -0.003 (-0.074-0.068) |
| $R_{intra}$ | 8 | **0.223** (0.161-0.284) | 11 | 0.199 (0.144-0.254) | 11 | 0.177 (0.105-0.248) | 12 | 0.208 (0.151-0.265) |
| $R_\Delta$ | 15 | 0.554 (0.507-0.602) | 12 | **0.586** (0.550-0.622) | 5 | 0.080 (-0.010-0.171) | 15 | 0.580 (0.532-0.628) |
| $D_{init}$ | 9 | -0.070 (-0.113—0.028) | 1 | -0.049 (-0.094--0.004) | 2 | -0.123 (-0.181--0.065) | 8 | **-0.012** (-0.059-0.034) |
| $D_{intra}$ | 11 | 0.302 (0.257-0.346) | 14 | **0.344** (0.287-0.401) | 12 | 0.134 (0.030-0.239) | 11 | 0.277 (0.233-0.322) |
| $D_\Delta$ | 14 | **0.500** (0.450-0.549) | 15 | 0.499 (0.452-0.546) | 15 | 0.180 (0.102-0.257) | 14 | 0.471 (0.415-0.528) |
| $R_{init} + R_{intra} + R_\Delta$ | 14 | 0.551 (0.507-0.594) | 14 | **0.643** (0.600-0.686) | 9 | 0.231 (0.118-0.343) | 10 | 0.536 (0.492-0.580) |
| $D_{init} + D_{intra} + D_\Delta$ | 14 | **0.563** (0.510-0.617) | 15 | 0.537 (0.483-0.591) | 15 | 0.143 (0.012-0.274) | 14 | 0.539 (0.475-0.603) |
| $R_{init} + R_{intra} + R_\Delta + D_{init} + D_{intra} + D_\Delta$ | 14 | 0.726 (0.689-0.763) | 14 | **0.743** (0.710-0.775) | 15 | 0.292 (0.172-0.411) | 14 | 0.728 (0.693-0.764) |

**Table 2.** Best RRMSE results for each scenario in the *X-cnt* criterion. The highest RRMSE across four kernels is highlighted in bold. 95% CIs are included in brackets.

|  | Linear | | RBF | | Polynomial | | Sigmoid | |
|---|---|---|---|---|---|---|---|---|
|  | Best # of selected features | RRMSE (95% CI) | Best # of selected features | RRMSE (95% CI) | Best # of selected features | RRMSE (95% CI) | Best # of selected features | RRMSE (95% CI) |
| $R_{init}$ | 13 | 0.113 (0.095-0.132) | 13 | **0.105** (0.090-0.121) | 4 | 0.124 (0.108-0.141) | 13 | 0.123 (0.105-0.142) |
| $R_{intra}$ | 7 | 0.112 (0.090-0.133) | 7 | 0.112 (0.090-0.134) | 8 | **0.106** (0.084-0.128) | 8 | 0.112 (0.091-0.134) |
| $R_\Delta$ | 11 | 0.048 (0.039-0.058) | 12 | **0.039** (0.035-0.044) | 10 | 0.090 (0.079-0.101) | 12 | 0.045 (0.037-0.053) |
| $D_{init}$ | 9 | 0.171 (0.147-0.195) | 15 | 0.151 (0.129-0.173) | 10 | 0.174 (0.148-0.200) | 9 | **0.148** (0.128-0.169) |
| $D_{intra}$ | 10 | 0.091 (0.076-0.106) | 15 | **0.076** (0.063-0.088) | 13 | 0.105 (0.085-0.124) | 10 | 0.093 (0.078-0.107) |
| $D_\Delta$ | 14 | 0.053 (0.044-0.063) | 15 | 0.052 (0.043-0.060) | 15 | 0.109 (0.092-0.125) | 14 | **0.051** (0.044-0.058) |
| $R_{init} + R_{intra} + R_\Delta$ | 10 | 0.045 (0.038-0.051) | 9 | **0.038** (0.032-0.044) | 10 | 0.075 (0.066-0.084) | 10 | 0.046 (0.039-0.052) |
| $D_{init} + D_{intra} + D_\Delta$ | 14 | 0.047 (0.038-0.056) | 14 | **0.045** (0.038-0.051) | 15 | 0.112 (0.087-0.137) | 14 | 0.046 (0.038-0.055) |
| $R_{init} + R_{intra} + R_\Delta + D_{init} + D_{intra} + D_\Delta$ | 14 | 0.024 (0.021-0.027) | 14 | **0.022** (0.020-0.025) | 15 | 0.068 (0.059-0.077) | 14 | 0.023 (0.020-0.026) |

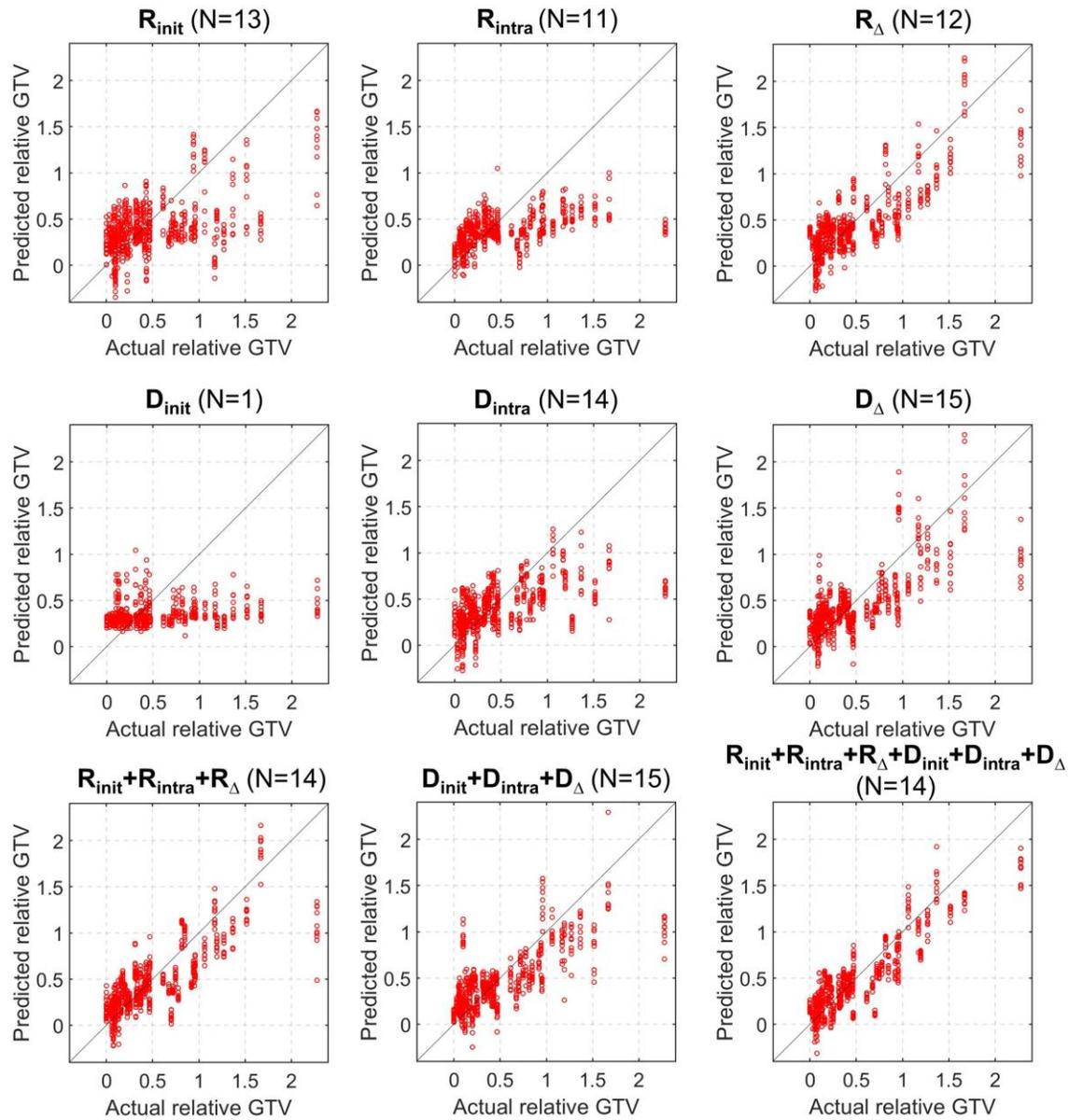

**Figure 5.** Scatter plots of actual GTV change (relative) versus predicted GTV change (relative) for nine scenarios ($X$-$cnt$ criterion, RBF kernel). The N number represents the total number of features with the best $R^2$ performance in each scenario.

## 3.4. Feature Correlation Analysis

The linear correlation coefficients of 15 selected features (using $X$-$cnt$ criterion) to the relative GTVs at follow up are shown in **Figure 6**, including the average of absolute coefficients accompanied with their standard deviation (SD). Multi-omic features show relatively stronger correlations compared to their corresponding individual-omic models in the upstream analysis. $\mathbf{R_{init}}+\mathbf{R_{intra}}+\mathbf{R_{\Delta}}+\mathbf{D_{init}}+\mathbf{D_{intra}}+\mathbf{D_{\Delta}}$ has the strongest correlation ($\overline{|r_i|}$ = 0.392) among nine scenarios.

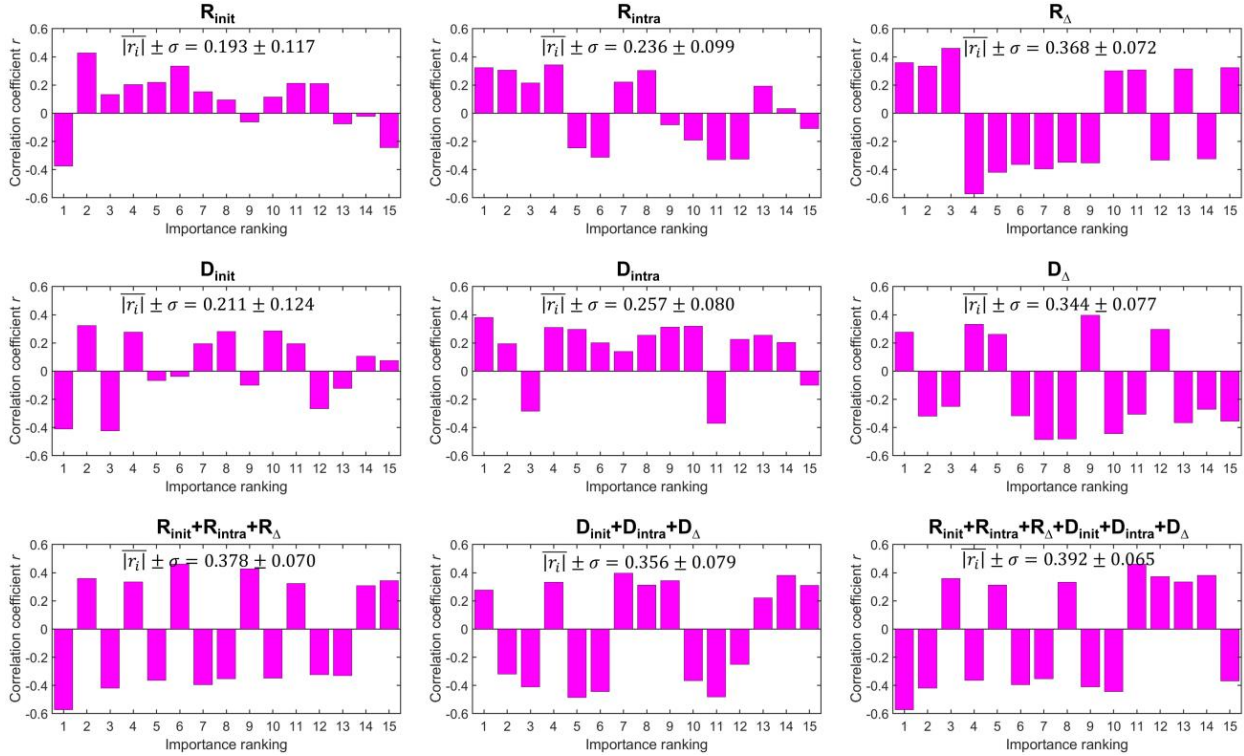

**Figure 6.** Linear correlation coefficients between 15 selected features (*X-cnt* criterion) and the relative GTVs at follow-up compared to the pre-treatment. The average of the 15 absolute coefficients accompanied with standard error is displayed in each panel.

**Figure 7** exhibits the correlation heatmap of the selected 15-feature sets (*X-cnt* criterion), showing the pairwise feature correlation coefficients, including the average of absolute coefficients accompanied with their SD. For $\mathbf{R_{init}}+\mathbf{R_{intra}}+\mathbf{R_{\Delta}}+\mathbf{D_{init}}+\mathbf{D_{intra}}+\mathbf{D_{\Delta}}$, the pairwise feature association is found to be moderate ($\overline{|r_i|} \pm \sigma$ = 0.150 ± 0.125) in comparison to others. This suggests that moderate correlations among selected features are more conducive to build a robust and accurate SVR model. Additionally, multicollinearity analysis among the 15 selected features assessed using the variance inflation factor (VIF), is shown in **Figure 8** and **Figure S9** (**Supplementary Material**). For the *X-cnt* criterion, all VIF values are below 5, indicating weak multicollinearity existing among the 15 features. In contrast, under the *X-abs* criterion, several scenarios exhibit VIF values exceeding 5, suggesting the presence of strong multicollinearity.

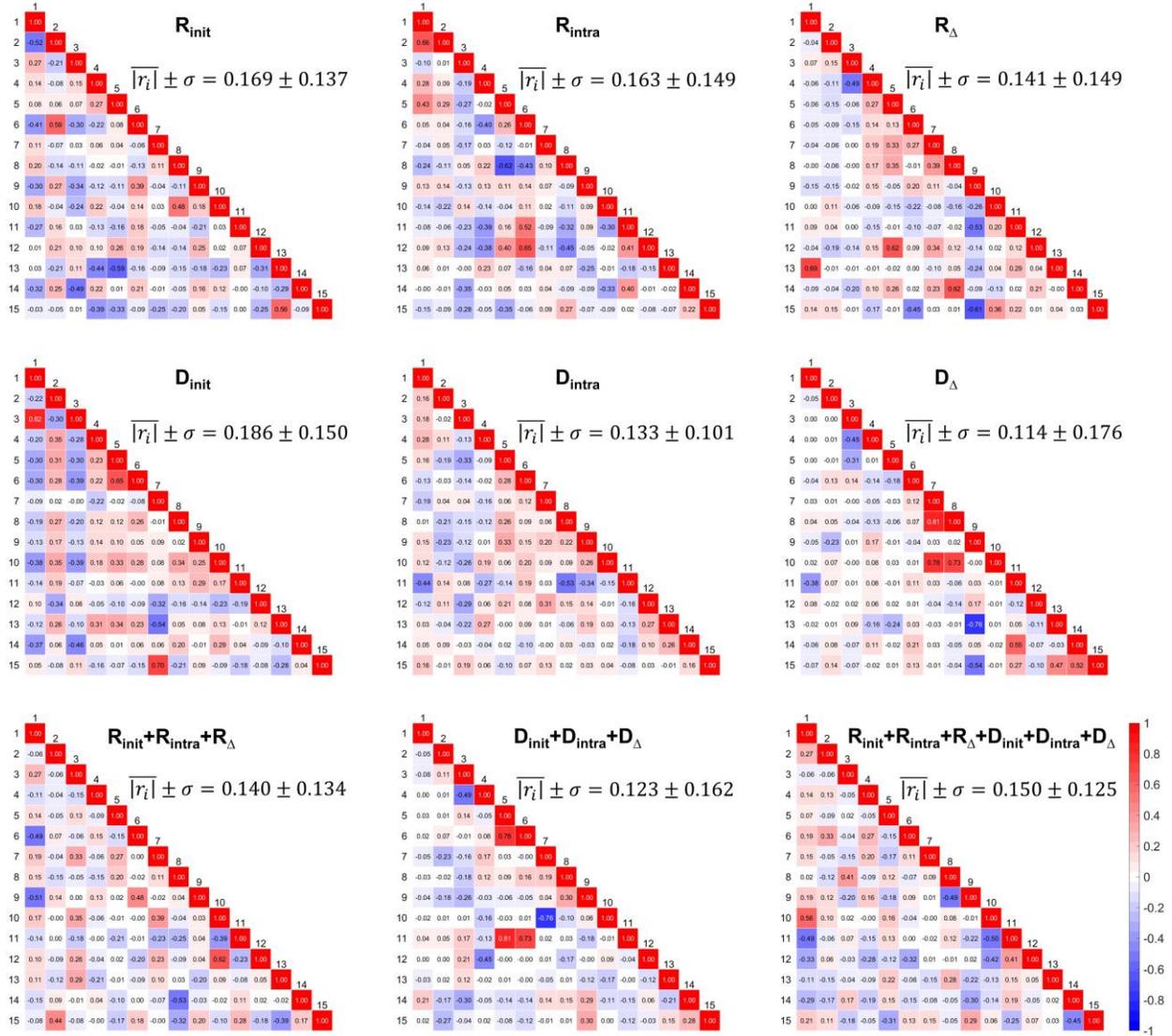

**Figure 7.** Pair-wise correlation heatmaps of features (*X-cnt* criterion). The average of the 15 absolute coefficients accompanied with standard error is displayed in each panel.

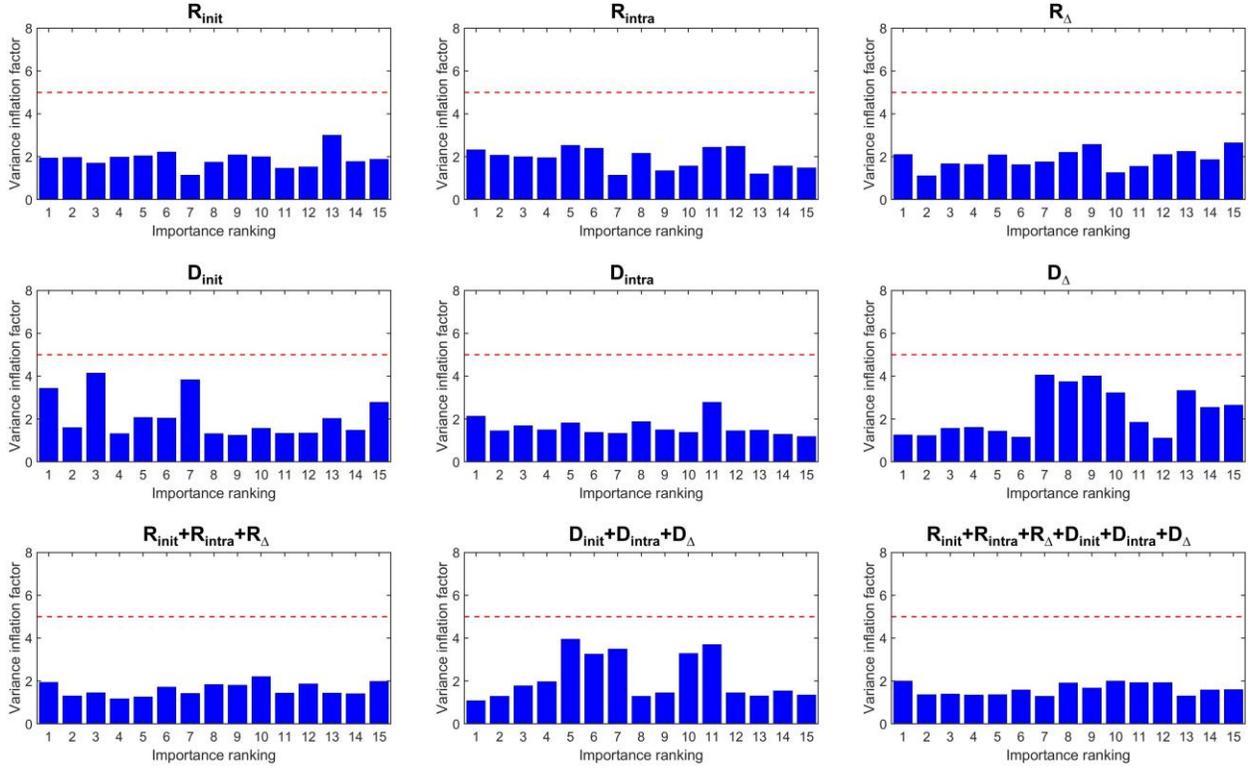

**Figure 8.** Multicollinearity analysis via variance inflation factors (VIFs) (*X-cnt* criterion).

## 3.5. Feature Effect Size Analysis

To quantitatively assess the capability of selected features to discriminate the dual groups split from 69 samples, Cohen's *d* was applied to express the feature effect size between two groups. **Figure 9** presents the effect size analysis of 15 features (*X-cnt*) with a group splitting threshold of 0.6 for relative GTVs, by dividing Cohen's *d* into four intervals (i.e., $d < 0.2$, $0.2 \leq d < 0.5$, $0.5 \leq d < 0.8$, and $d \geq 0.8$). Such partition was chosen according to the guidelines to interpret the magnitude of difference with 0.2 being "small", 0.5 being "medium", and 0.8 being "large"(Goulet-Pelletier and Cousineau, 2018). It is found that multi-omic feature sets exhibit higher effect sizes (more greed blocks) compared with individual-omic ones, except for $\mathbf{R}_\Delta$. However, for the average Cohen's *d* values, the largest value across nine scenarios is 0.69 achieved by $\mathbf{R}_{init}+\mathbf{R}_{intra}+\mathbf{R}_\Delta+\mathbf{D}_{init}+\mathbf{D}_{intra}+\mathbf{D}_\Delta$, which reveals its most discriminative power between the two groups. Other splitting thresholds (i.e., 0.2, 0.4, and 0.8) are shown in **Figure S10-S12,** demonstrating a similar trend.

| Importance rankings | $R_{int}$ | | $R_{intra}$ | | $R_\Delta$ | | $D_{init}$ | | $D_{intra}$ | | $D_\Delta$ | | $R_{init}+R_{intra}+R_\Delta$ | | $D_{init}+D_{intra}+D_\Delta$ | | $R_{init}+R_{intra}+R_\Delta$ + $D_{init}+D_{intra}+D_\Delta$ | |
|---|---|---|---|---|---|---|---|---|---|---|---|---|---|---|---|---|---|---|
| | Distribution* | Cohen's $d^\&$ | Distribution* | Cohen's $d^\&$ | Distribution* | Cohen's $d^\&$ | Distribution* | Cohen's $d^\&$ | Distribution* | Cohen's $d^\&$ | Distribution* | Cohen's $d^\&$ | Distribution* | Cohen's $d^\&$ | Distribution* | Cohen's $d^\&$ | Distribution* | Cohen's $d^\&$ |
| 1 | | 0.79 | | 0.59 | | 0.66 | | 0.80 | | 0.51 | | 0.44 | | 0.64 | | 0.44 | | 0.64 |
| 2 | | 0.46 | | 0.42 | | 0.81 | | 0.57 | | 0.33 | | 0.55 | | 0.66 | | 0.55 | | 0.69 |
| 3 | | 0.25 | | 0.33 | | 0.63 | | 0.82 | | 0.73 | | 0.74 | | 0.69 | | 0.80 | | 0.66 |
| 4 | | 0.41 | | 0.66 | | 0.64 | | 0.51 | | 0.65 | | 0.92 | | 0.81 | | 0.92 | | 0.76 |
| 5 | | 0.42 | | 0.33 | | 0.69 | | 0.12 | | 0.78 | | 0.51 | | 0.76 | | 0.45 | | 1.04 |
| 6 | | 0.52 | | 0.54 | | 0.76 | | 0.06 | | 0.38 | | 0.70 | | 0.63 | | 0.45 | | 0.64 |
| 7 | | 0.34 | | 0.46 | | 0.64 | | 0.50 | | 0.16 | | 0.45 | | 0.64 | | 0.64 | | 0.65 |
| 8 | | 0.16 | | 0.39 | | 0.56 | | 0.50 | | 0.40 | | 0.56 | | 0.65 | | 1.04 | | 0.92 |
| 9 | | 0.23 | | 0.18 | | 0.65 | | 0.18 | | 1.04 | | 0.64 | | 0.46 | | 0.79 | | 0.80 |
| 10 | | 0.09 | | 0.43 | | 0.67 | | 0.61 | | 0.58 | | 0.45 | | 0.56 | | 0.59 | | 0.45 |
| 11 | | 0.33 | | 0.49 | | 0.66 | | 0.59 | | 0.61 | | 0.63 | | 0.59 | | 0.56 | | 0.63 |
| 12 | | 0.34 | | 0.52 | | 0.68 | | 0.45 | | 0.47 | | 0.73 | | 0.87 | | 0.74 | | 0.42 |
| 13 | | 0.19 | | 0.44 | | 0.56 | | 0.30 | | 0.53 | | 0.59 | | 0.49 | | 0.58 | | 0.81 |
| 14 | | 0.02 | | 0.16 | | 0.87 | | 0.24 | | 0.18 | | 0.54 | | 0.66 | | 0.51 | | 0.51 |
| 15 | | 0.58 | | 0.32 | | 0.52 | | 0.15 | | 0.33 | | 0.58 | | 0.66 | | 0.65 | | 0.71 |
| Averaged Cohen's $d$ | | 0.34 | | 0.42 | | 0.67 | | 0.43 | | 0.51 | | 0.60 | | 0.65 | | 0.65 | | **0.69** |

\* Relative GTV < 0.6; Relative GTV ≥ 0.6.
\& Effect size levels: $d < 0.2$; $0.2 \leq d < 0.5$; $0.5 \leq d < 0.8$; $d \geq 0.8$.

**Figure 9.** Feature effect size analysis of selected 15 features (*X-cnt* criterion) with a group splitting threshold of 0.6. Each feature contains distribution of feature values and corresponding Cohen's *d* value. The color coding for Cohen's *d* values indicates the varying effect size levels.

## 4. Discussion

In this study we demonstrate the feasibility of using radiomics and dosiomics to predict GTV change, targeting for the PULSAR treatment. PULSAR, characterized by two treatment courses separated by a protracted interval, enables more meaningful adaptation based on patient-specific tumor responses. The acquisition of multiple MRI images (e.g., two scans at this work) reveals tumor morphologic and functional dynamics. Therefore, both radiomic and dosiomic features at each time point, along with their delta-omics, are available for outcome prediction. Unlike classification models, regression models offer detailed, quantitative insights into tumor changes. Tumor progression is continuous, and regression captures subtle variations in size—enabling estimation of change magnitude (e.g., percent increase or decrease) rather than binary outcomes. This enhanced precision can help support more informed treatment decisions, such as dose boosting or systemic treatment. Moreover, clinicians can define custom thresholds for categorization, and when monitoring tumors over time, regression better quantifies dynamic trends of treatment response.

For the first time, we shift from a classification paradigm to a regression framework for modeling GTV change. In the best-case scenario, the top-performing model achieved an $R^2$ of 0.743 and a RRMSE of 0.022 (**Figure 4**). This means that approximately 74% of the variance in tumor volume change can be explained by the omics-based regression model, while the remaining 26% likely stems from factors not captured by the features or from intrinsic biological and imaging noise. To validate the strength of this model and ensure interpretability, we employed a series of rigorous statistical analyses. These included pairwise feature correlation, correlation between features and the GTV label, VIF assessments to check multicollinearity, and effect size visualization using Cohen's *d* values. These metrics collectively provide insights into both feature robustness and relevance. Moreover, the performance reported here is not from a single run but rather derived from robust validation using repeated cross-validation—specifically, 10 repetitions of 5-fold cross-validation. This enhances our confidence in the model's generalizability, although the possibility of overfitting cannot be entirely ruled out. We anticipate that being able to explain over 70% of the variance in a continuous outcome like GTV change can significantly assist timely treatment adjustment and patient-specific decision-making. Unlike classification, which yields only categorical labels, regression allows for fine-grained, interpretable outputs such as percentage increase or decrease in tumor volume. Nonetheless, it is important to acknowledge the complexity of modeling tumor response. Tumor progression or regression is governed by a confluence of factors—including radiomic texture, spatial dose deposition, and underlying biological heterogeneity—all of which may interact in nonlinear or multimodal ways. While these initial findings are promising, validation in a larger, independent PULSAR cohort is essential before drawing definitive clinical conclusions. Until then, cautious optimism is appropriate.

Compared to individual-omics, multi-omics has already proved to be a more promising strategy for improving tumor treatment response prediction(Zhang *et al.*, 2023; Wang *et al.*, 2023; Zheng *et al.*, 2023). Its superiority found in our study can be attributed to three main aspects. First, multi-omics integrates complementary dataset, reducing bias associated with any single source and thereby enhancing predictive accuracy. Second, multi-omics has potential to bridge potential interactions among single-omic features. To elaborate, radiomics extracts quantitative features from MRI and captures tumor shape, texture, intensity, and spatial heterogeneity. These features can reflect underlying tumor biology, such as aggressiveness, hypoxia, or response potential. Dosiomics analyzes the spatial dose distribution from radiation therapy plans, characterizing how radiation is delivered across the tumor and surrounding tissue, capturing information such as dose gradients, hot/cold spots, and conformity. Radiomics describes "what the tumor looks like," while dosiomics shows "how it's being treated." When combined, they provide a more holistic view of the treatment context—linking tumor characteristics with how effectively radiation targets them. For example, a region of high heterogeneity may receive a lower dose due to proximity to sensitive organs — that interaction can be critical for outcome. Finally, while features from individual time points provide valuable information, delta-omics—which captures changes in features over time—often yields superior predictive performance(Nardone *et al.*, 2021; Nardone *et al.*, 2024). Integrating both static and delta-omic features enables the construction of more robust and accurate prediction models, particularly for PULSAR.

After selecting a subset of features, ranking them helps identify which features contribute most to model predictions. When combining multi-omics (e.g., radiomics + dosiomics), feature ranking can inform fusion strategies based on feature importance. In our study, feature ranking is based on nonzero weights obtained from Lasso, using *X-abs* and *X-cnt*. The *X-abs* criterion determined by mean absolute coefficients across multiple iterations tends to select the most influential features, while the *X-cnt* criterion prefers stable and frequently important features. The relatively inferior performance generated by *X-abs*, suggests that it might over-prioritize features. For instance, despite *X-abs* criterion seeks for high correlation coefficients with respect to GTVs, the aggregate relevance of 15 selected features for *X-abs* is weaker than that for *X-cnt* in the multi-omic scenario (see $\mathbf{R_{init}}+\mathbf{R_{intra}}+\mathbf{R_\Delta}+\mathbf{D_{init}}+\mathbf{D_{intra}}+\mathbf{D_\Delta}$ in **Figure 6** and **Figure S9**). Moreover, the 15 correlation coefficients selected using the *X-cnt* criterion exhibit a more balanced distribution of positive and negative signs compared to those selected by the *X-abs* method (see **Figure 6** and **Figure S8** for their relationships with GTVs). This observation suggests that the *X-cnt* criterion may provide a more effective feature selection strategy, despite the fact that *X-abs* has been more commonly used in previous studies(Ghosh *et al.*, 2021; Demircioğlu, 2022).

Radiomic interpretability remains a major challenge. While radiomics can uncover hidden imaging patterns, linking these features to clear biological or clinical meanings is not straightforward. To explore their potential relevance, we briefly examine the top three selected features listed at **R$_{init}$+R$_{intra}$+R$_\Delta$+D$_{init}$+D$_{intra}$+D$_\Delta$** in **Figure 2** (*Wavelet LHH glrlm LongRunLowGrayLevelEmphasis*, *Original glrlm LowGrayLevelRunEmphasis*, and *Wavelet HLH glcm Autocorrelation*). The *Wavelet LHH glrlm LongRunLowGrayLevelEmphasis* quantifies the presence of extended runs of low-intensity pixels following wavelet decomposition. High values of this feature correspond to large, homogeneous regions of low density, typically associated with necrosis, edema, or hypoxic areas within the tumor. In contrast, the *Original glrlm LowGrayLevelRunEmphasis*, extracted from the unfiltered image, captures the overall contribution of low-gray-level runs irrespective of run length. Elevated values may indicate a predominance of darker tissue intensities, likely linked to poor vascularization or tumor heterogeneity. Lastly, the *Wavelet HLH glcm Autocorrelation* measures the spatial uniformity and repetition of gray levels in the wavelet-transformed image. A high autocorrelation value suggests a more regular and structured texture, potentially associated with well-structured tumors, whereas low values may point to heterogeneous or invasive phenotypes. Altogether, these features provide a multi-scale view of tumor architecture and heterogeneity. Future research incorporating histopathological correlation and biological validation is still crucial to confirm these mechanistic interpretations and validate their prognostic or predictive value.

Since PULSAR is a novel treatment approach recently established at our institution, we currently have a limited cohort of only 69 cases. These cases exhibit an imbalanced distribution of GTV (gross tumor volume) changes. For example, when applying an 80% threshold to split responses, 55 cases are classified as responders (GTV change >80%), and 14 as non-responders (GTV change <80%). This class imbalance has posed challenges in our previous classification studies. For regression tasks, we believe such imbalance is less of a concern. Nevertheless, small sample sizes remain a major challenge, especially for high-dimensional regression, potentially leading to overfitting and poor generalizability. To help mitigate this, we limited the number of features below 15, following a commonly recommended feature-to-sample ratio of 1:4(MacCallum *et al.*, 2001). Moreover, using the *X-cnt* criterion, we selected useful features that consistently appeared across samples, helping to reduce noise-driven or spurious correlations. We believe it is also possible to further reduce the feature count to 7 while still maintaining performance, as evidenced by an R$^2$ of 0.60 (**Figure 4**). In practice, classification tasks are often more prone to overfitting, especially when working with imbalanced classes, because it relies on discrete boundaries and is highly sensitive to small data variations. In contrast, regression models predict continuous values and are less susceptible to overfitting, as they must learn actual data relationships rather than just separating classes (**Figure 5**). While our current regression model shows promise, we have to validate it on an independent dataset to rigorously assess potential overfitting once more PULSAR patients become available.

Encouragingly, $R^2$ does not continue to increase with the number of features (**Figure 4**). This suggests two possibilities. First, the upper limit of predictability for our regression framework has been reached, implying that there are other confounding factors, such as patient-specific characteristics or tumor microenvironment that cannot be reflected by radiomic or dosimetric data alone. Second, it could mean that the model is approaching the optimal level of complexity, with around 10 multi-omic features being sufficient for accurate GTV regression.

Below are some comments on the comparison of the four kernels presented in **Table 1** and **Table 2**. In the multi-omic scenario, the difference between the RBF and linear kernels is minimal ($R^2$: 0.743 vs. 0.726, RRMSE: 0.022 vs. 0.024). This suggests that the relationship between features and GTV change is largely linear, so that the advantage of using the RBF kernel is limited. It is important to note that the two parameters in the SVR-RBF have been carefully fine-tuned in our study (see **Supplementary Material**) to mitigate overfitting, by controlling the penalty for predictions outside the margin and the kernel parameter. The reason why the RBF kernel slightly outperform the linear kernel might be due to its handling multicollinearity, which the linear kernel does not inherently resolve. Linear models can assign unstable or exaggerated weights to highly correlated features, which hurts generalization. The RBF kernel, by projecting data into a higher-dimensional space using nonlinear transformations, can reduce the impact of correlated features by spreading them apart in the new space. This enables the model to extract more robust and discriminative features, improving prediction accuracy while minimizing redundancy. The reason why the polynomial kernel performed poorly ($R^2$: 0.292, RRMSE: 0.068), suggesting that the added complexity from polynomial transformations is harmful in this setting.

Beyond the limited cohort size, our current study has several other limitations that warrant further investigation. First, relative GTVs are heavily clustered below 0.5, so predictions for larger volumes (those non-response lesions) exhibit larger error (**Figure 5**). Second, although we used SVR, additional regression methods, such as ordinary least squares(James *et al.*, 2023), Gaussian processes regression(Schulz *et al.*, 2018), artificial neural networks(Specht, 1991; Rodriguez-Galiano *et al.*, 2015), should be evaluated. Third, lesion targets depend on manual or semi-automated segmentation, so even small segmentation or calculation errors can significantly skew results. Finally, as more follow-up data accrue, future work should extend beyond tumor volume to incorporate long-term clinical endpoints including overall survival and disease progression.

## 5. Conclusions

In summary, we employed radiomics and dosiomics to develop support vector regression (SVR) models for estimating follow-up GTV within the PULSAR treatment framework. Unlike the common use of radiomics and dosiomics in classification tasks, regression models provide continuous value predictions, allowing for more nuanced quantification and interpretability. The frequency-based feature selection criterion proves more effective than the weight-based approach. Multi-omic models—combining radiomics, dosiomics, and their delta forms—outperform single-omic counterparts. Among the four SVR kernel functions tested (linear, RBF, polynomial, and sigmoid), the RBF kernel yields superior results. As our PULSAR cohort expands, we anticipate continued improvements in model accuracy and robustness, enhancing the potential of the regression model as a quantitative and personalized tool to support patient selection and treatment adaptation in PULSAR.

## Acknowledgements

We express our gratitude to Drs. Hua-Chieh Shao, Guoping Xu, Bowen Jing, and Ying Luo for their insightful discussions to this work.

## Author contributions

Y.Y. and H.P. conceived and designed the study. H.P. and R.T. verified the underlying raw data. Y.Y. developed, trained, and tested the prediction models. Y.Y., S.J. and H.P. contributed to the statistical analysis and interpretation of data. All the authors contributed to manuscript preparation and approved the final manuscript.

## Conflict of interest

The authors have no relevant conflicts of interest to disclose.

## Data and code availability statement

All data that support the findings of this study are included within the article and the supplementary files. The codes in the analyses are made available publicly as a GitHub repository at https://github.com/yajunyu9999/PULSAR-Regression.

# REFERENCES

Awad M and Khanna R 2015 *Efficient Learning Machines: Theories, Concepts, and Applications for Engineers and System Designers,* ed M Awad and R Khanna (Berkeley, CA: Apress) pp 67-80

Burnet N G, Thomas S J, Burton K E and Jefferies S J 2004 Defining the tumour and target volumes for radiotherapy *Cancer Imaging* **4** 153-61

Demaria S, Guha C, Schoenfeld J, Morris Z, Monjazeb A, Sikora A, Crittenden M, Shiao S, Khleif S, Gupta S, Formenti S C, Vikram B, Coleman C N and Ahmed M M 2021 Radiation dose and fraction in immunotherapy: one-size regimen does not fit all settings, so how does one choose? *J Immunother Cancer* **9**

Demircioğlu A 2022 Benchmarking feature selection methods in radiomics *Investigative radiology* **57** 433-43

Dohopolski M, Schmitt L G, Anand S, Zhang H, Stojadinovic S, Youssef M, Shaikh N, Patel T, Patel A, Barnett S, Lee D S, Ahn C, Lee M, Timmerman R, Peng H, Cai X, Dan T and Wardak Z 2024 Exploratory Evaluation of Personalized Ultrafractionated Stereotactic Adaptive Radiation Therapy (PULSAR) With Central Nervous System-Active Drugs in Brain Metastases Treatment *International Journal of Radiation Oncology\*Biology\*Physics*

Elshafeey N, Kotrotsou A, Hassan A, Elshafei N, Hassan I, Ahmed S, Abrol S, Agarwal A, El Salek K, Bergamaschi S, Acharya J, Moron F E, Law M, Fuller G N, Huse J T, Zinn P O and Colen R R 2019 Multicenter study demonstrates radiomic features derived from magnetic resonance perfusion images identify pseudoprogression in glioblastoma *Nature Communications* **10** 3170

Fakiris A J, McGarry R C, Yiannoutsos C T, Papiez L, Williams M, Henderson M A and Timmerman R 2009 Stereotactic Body Radiation Therapy for Early-Stage Non–Small-Cell Lung Carcinoma: Four-Year Results of a Prospective Phase II Study *International Journal of Radiation Oncology\*Biology\*Physics* **75** 677-82

Ghosh P, Azam S, Jonkman M, Karim A, Shamrat F J M, Ignatious E, Shultana S, Beeravolu A R and De Boer F 2021 Efficient prediction of cardiovascular disease using machine learning algorithms with relief and LASSO feature selection techniques *IEEE Access* **9** 19304-26

Goulet-Pelletier J-C and Cousineau D 2018 A review of effect sizes and their confidence intervals, Part I: The Cohen's d family *The Quantitative Methods for Psychology* **14** 242-65

James G, Witten D, Hastie T, Tibshirani R and Taylor J 2023 *An introduction to statistical learning: With applications in python*: Springer) pp 69-134

Jiang Y, Zhang Z, Wang W, Huang W, Chen C, Xi S, Ahmad M U, Ren Y, Sang S, Xie J, Wang J-Y, Xiong W, Li T, Han Z, Yuan Q, Xu Y, Xing L, Poultsides G A, Li G and Li R 2023 Biology-guided deep learning predicts prognosis and cancer immunotherapy response *Nature Communications* **14** 5135

Kumar V, Gu Y, Basu S, Berglund A, Eschrich S A, Schabath M B, Forster K, Aerts H J, Dekker A and Fenstermacher D 2012 Radiomics: the process and the challenges *Magnetic resonance imaging* **30** 1234-48

Lambin P, Leijenaar R T, Deist T M, Peerlings J, De Jong E E, Van Timmeren J, Sanduleanu S, Larue R T, Even A J and Jochems A 2017 Radiomics: the bridge between medical imaging and personalized medicine *Nature reviews Clinical oncology* **14** 749-62

Liang B, Yan H, Tian Y, Chen X, Yan L, Zhang T, Zhou Z, Wang L and Dai J 2019 Dosiomics: Extracting 3D Spatial Features From Dose Distribution to Predict Incidence of Radiation Pneumonitis *Frontiers in Oncology* **9**

Lo S S, Fakiris A J, Chang E L, Mayr N A, Wang J Z, Papiez L, Teh B S, McGarry R C, Cardenes H R and Timmerman R D 2010 Stereotactic body radiation therapy: a novel treatment modality *Nature reviews Clinical oncology* **7** 44-54

MacCallum R C, F. W K, J. P K and and Hong S 2001 Sample Size in Factor Analysis: The Role of Model Error *Multivariate Behavioral Research* **36** 611-37

Miljanic M, Montalvo S, Aliru M, Song T, Leon-Camarena M, Innella K, Vujovic D, Komaki R and Iyengar P 2022 The Evolving Interplay of SBRT and the Immune System, along with Future Directions in the Field *Cancers* **14** 4530

Moore C, Hsu C-C, Chen W-M, Chen B P C, Han C, Story M, Aguilera T, Pop L M, Hannan R, Fu Y-X, Saha D and Timmerman R 2021 Personalized Ultrafractionated Stereotactic Adaptive Radiotherapy (PULSAR) in Preclinical Models Enhances Single-Agent Immune Checkpoint Blockade *International Journal of Radiation Oncology\*Biology\*Physics* **110** 1306-16

Mu W, Jiang L, Zhang J, Shi Y, Gray J E, Tunali I, Gao C, Sun Y, Tian J, Zhao X, Sun X, Gillies R J and Schabath M B 2020 Non-invasive decision support for NSCLC treatment using PET/CT radiomics *Nature Communications* **11** 5228

Murakami Y, Soyano T, Kozuka T, Ushijima M, Koizumi Y, Miyauchi H, Kaneko M, Nakano M, Kamima T, Hashimoto T, Yoshioka Y and Oguchi M 2022 Dose-Based Radiomic Analysis (Dosiomics) for Intensity Modulated Radiation Therapy in Patients With Prostate Cancer: Correlation Between Planned Dose Distribution and Biochemical Failure *International Journal of Radiation Oncology\*Biology\*Physics* **112** 247-59

Nardone V, Reginelli A, Grassi R, Boldrini L, Vacca G, D'Ippolito E, Annunziata S, Farchione A, Belfiore M P and Desideri I 2021 Delta radiomics: A systematic review *La radiologia medica* **126** 1571-83

Nardone V, Reginelli A, Rubini D, Gagliardi F, Del Tufo S, Belfiore M P, Boldrini L, Desideri I and Cappabianca S 2024 Delta radiomics: an updated systematic review *La radiologia medica*  1-18

Nasief H, Zheng C, Schott D, Hall W, Tsai S, Erickson B and Allen Li X 2019 A machine learning based delta-radiomics process for early prediction of treatment response of pancreatic cancer *NPJ precision oncology* **3** 25

Peng H, Deng J, Jiang S and Timmerman R 2024a Rethinking the potential role of dose painting in personalized ultra-fractionated stereotactic adaptive radiotherapy *Frontiers in Oncology* **14**

Peng H, Moore C, Zhang Y, Saha D, Jiang S and Timmerman R 2024b An AI-based approach for modeling the synergy between radiotherapy and immunotherapy *Scientific Reports* **14** 8250

Plachouris D, Eleftheriadis V, Nanos T, Papathanasiou N, Sarrut D, Papadimitroulas P, Savvidis G, Vergnaud L, Salvadori J, Imperiale A, Visvikis D, Hazle J D and Kagadis G C 2023 A radiomic- and dosiomic-based machine learning regression model for pretreatment planning in 177Lu-DOTATATE therapy *Medical Physics* **50** 7222-35

Rodriguez-Galiano V, Sanchez-Castillo M, Chica-Olmo M and Chica-Rivas M 2015 Machine learning predictive models for mineral prospectivity: An evaluation of neural networks, random forest, regression trees and support vector machines *Ore Geology Reviews* **71** 804-18

Schulz E, Speekenbrink M and Krause A 2018 A tutorial on Gaussian process regression: Modelling, exploring, and exploiting functions *Journal of Mathematical Psychology* **85** 1-16

Slawson R G, Salazar O M, Poussin-Rosillo H, Amin P P, Strohl R and Sewchand W 1988 Once-a-week vs conventional daily radiation treatment for lung cancer: Final report *International Journal of Radiation Oncology*Biology*Physics* **15** 61-8

Smola A J and Schölkopf B 2004 A tutorial on support vector regression *Statistics and Computing* **14** 199-222

Specht D F 1991 A general regression neural network *IEEE transactions on neural networks* **2** 568-76

Teruel J R, Heldahl M G, Goa P E, Pickles M, Lundgren S, Bathen T F and Gibbs P 2014 Dynamic contrast-enhanced MRI texture analysis for pretreatment prediction of clinical and pathological response to neoadjuvant chemotherapy in patients with locally advanced breast cancer *NMR in Biomedicine* **27** 887-96

Thompson C G, Kim R S, Aloe A M and Becker B J 2017 Extracting the Variance Inflation Factor and Other Multicollinearity Diagnostics from Typical Regression Results *Basic and Applied Social Psychology* **39** 81-90

Tibshirani R 1996 Regression shrinkage and selection via the lasso *Journal of the Royal Statistical Society Series B: Statistical Methodology* **58** 267-88

Van Griethuysen J J, Fedorov A, Parmar C, Hosny A, Aucoin N, Narayan V, Beets-Tan R G, Fillion-Robin J-C, Pieper S and Aerts H J 2017 Computational radiomics system to decode the radiographic phenotype *Cancer research* **77** e104-e7

Wang B, Liu J, Zhang X, Wang Z, Cao Z, Lu L, Lv W, Wang A, Li S, Wu X and Dong X 2023 Prognostic value of 18F-FDG PET/CT-based radiomics combining dosiomics and dose volume histogram for head and neck cancer *EJNMMI Research* **13** 14

Wang K, Zhao J, Duan J, Feng C, Li Y, Li L and Yuan S 2024 Radiomic and dosimetric parameter-based nomogram predicts radiation esophagitis in patients with non-small cell lung cancer undergoing combined immunotherapy and radiotherapy *Frontiers in Oncology* **14**

Yu Y, Jiang S, Timmerman R and Peng H 2025 Leveraging Compressed Sensing and Radiomics for Robust Feature Selection for Outcome Prediction in Personalized Ultra-Fractionated Stereotactic Adaptive Radiotherapy *Advanced Intelligent Systems* **n/a** 2500116

Zhang H, Dohopolski M, Stojadinovic S, Schmitt L G, Anand S, Kim H, Pompos A, Godley A, Jiang S, Dan T, Wardak Z, Timmerman R and Peng H 2024 Multiomics-Based Outcome Prediction in Personalized Ultra-Fractionated Stereotactic Adaptive Radiotherapy (PULSAR) *Cancers* **16** 3425

Zhang Z, Wang Z, Yan M, Yu J, Dekker A, Zhao L and Wee L 2023 Radiomics and Dosiomics Signature From Whole Lung Predicts Radiation Pneumonitis: A Model Development Study With Prospective External Validation and Decision-curve Analysis *International Journal of Radiation Oncology\*Biology\*Physics* **115** 746-58

Zheng X, Guo W, Wang Y, Zhang J, Zhang Y, Cheng C, Teng X, Lam S, Zhou T, Ma Z, Liu R, Wu H, Ge H, Cai J and Li B 2023 Multi-omics to predict acute radiation esophagitis in patients with lung cancer treated with intensity-modulated radiation therapy *European Journal of Medical Research* **28** 126

Zwanenburg A, Leger S, Vallières M and Löck S 2016 Image biomarker standardisation initiative *arXiv preprint arXiv:1612.07003*